\documentclass[runningheads]{llncs}
\usepackage[T1]{fontenc}

\usepackage{graphicx}

\begin{document}

\title{Inaugural Lecture\\ Machine learning: progress and prospects}
\author{Alexander Gammerman}
\authorrunning{A. Gammerman}

\institute{Department of Computer Science, \\ Royal Holloway University of London, TW20 0EX, UK \\
\email{A.Gammerman@rhul.ac.uk}\\
\url{https://cml.rhul.ac.uk/people/alex/} }
\maketitle            
\begin{abstract}
This Inaugural Lecture was given at Royal Holloway University of London in 1996. It covers an introduction to machine learning and describes various theoretical advances and practical projects in the field.
The Lecture here is presented in its original format, but a few remarks have been added in 2025 to reflect recent developments, and the list of references has been updated to enhance the convenience and accuracy for readers.

\end{abstract}

\section{Introduction}

My subject in this lecture is machine learning: how to write {\bf algorithms}
and {\bf programs} that learn.  In the spirit of our time, just before I
started to write this lecture, I searched the Internet to see whether there
were any entries under ``Machine Learning''.  I expected to find several
hundred papers and other documents written on the subject, but found the
astonishing total of 400,000 documents written on Machine Learning and
accessible on the Internet in November 1996.  Surely, no-one can read that
many documents, and indeed probably 99\% of them are some trivial programs, or
re-inventing the wheel.  Perhaps it would be worth writing a machine learning
program to analyse this set of documents.

So I may be trying to do an impossible task: to review the progress in this
field.  I did not read one tenth, or one hundredth or even one
thousandth of the available papers, and inevitably this talk will be a very
personal view on the subject.

When did machine learning start?  Maybe a good starting point is 1949 when
Claude Shannon suggested a learning algorithm for chess playing programs.  Or
maybe we should go back to the 1930s when Ronald Fisher developed discriminant
analysis -- a type of learning where the problem is to construct a decision
rule that separates two types of vector.  Or could it be the 18th century when
David Hume discussed the idea of induction?  Or the 14th century when William
of Ockham formulated the principle of ``simplicity'' known as ``Ockham's
razor''?  (Ockham, by the way, is a small village not far from Royal
Holloway).  Or it may be that, like almost everything else in western
civilisation and culture, the origin of these ideas lies in the Mediterranean?
After all, it was Aristotle who said that ``we {\bf learn} some things only by
doing things''.

I would like, however, to start from the middle of this century -- the
computers have just arrived and, perhaps, this topic (ML) is as old as
computer science.  In 1950, Alan Turing had just published one of his best
papers (apart from his mathematical papers) entitled ``Computing Machinery and
Intelligence''~\cite{Turing}.  In order to make the machine intelligent (in some
sense) he suggested that the machine might be programmed to simulate a child's
brain, then equipped with a {\bf learning program} and taught like a child.
This brings out the point that {\bf thinking} ({\bf intelligence}) is closely
connected with {\bf learning}.

\begin{figure}[h]
	\centering
\includegraphics[width=0.6\textwidth, angle=270]{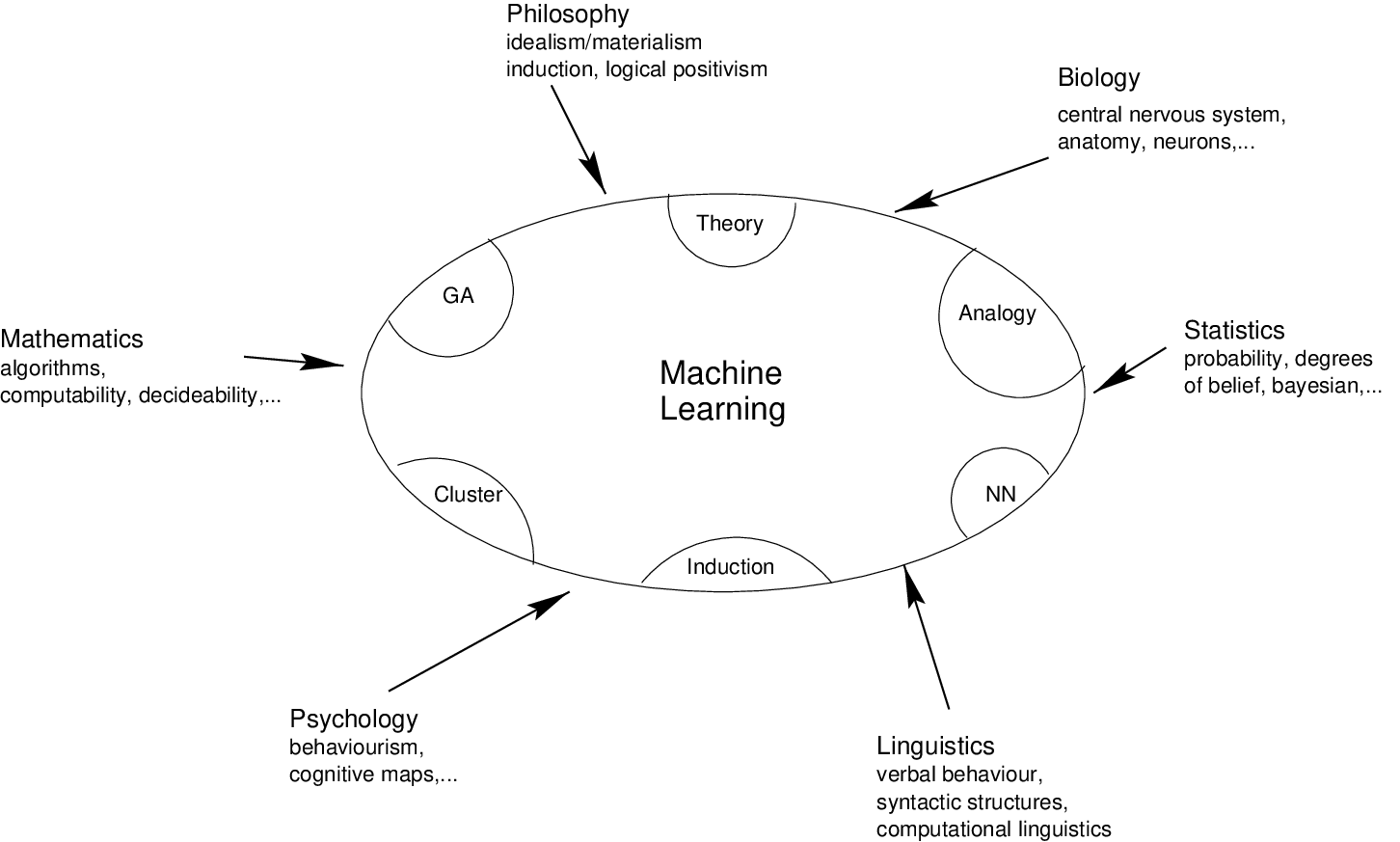}
\caption{Machine Learning} \label{fig1}
\end{figure}

The field of machine learning has been greatly influenced by other disciplines
and the subject is in itself not a very homogeneous discipline but includes
separate, overlapping subfields.  There are many parallel lines of research in
ML: inductive learning, neural networks (NN), clustering, learning by analogy,
genetic algorithms (GA), and theories of learning - they are all part of 
the more general field of machine learning - see Figure~\ref{fig1}.

Let me start by showing you several examples of achievements in machine
learning.

The first example is a system called ALVINN~\cite{Pom} 
which learns to steer a vehicle
along a motorway by observing the performance of a human driver.  The results
of the training are very impressive -- ALVINN has driven at speeds up to 70
mph for a distance of up to 90 miles on a motorway\footnote{After 30 years of development in the field, by 2025, the self-driving cars reached a commercial pilot stage.}.

Another example is the chess playing program.  Let us compare the ratings of
human and machine chess champions.  Back in 1960 the World Champion was Botvinnik
and he had a rating of 2,616, while the best machine in 1965 had a rating of
1,400.  By the year 1992 Kasparov, the current World Champion, had a rating of
2,805 while the best computer in 1994, Deep Thought 2, had a rating of 2,600
approx.  For comparison, a county player has a rating in the region of 2,100,
an international master of 2,400, and an international grandmaster of 2,500.
During the last 30 years or so, the performance of these programs has improved
enormously, and perhaps one can extrapolate this trend further\footnote{In 1997, just the following year after this Lecture was delivered,
IBM’s {\it Deep Blue} beat the champion Garry Kasparov. Between the 2000s - 2010s, new programs like {\it Fritz and Stockfish} were developed and they became accessible on consumer hardware. The introduction of deep learning has further advanced the development of chess-playing programs, such as {\it AlphaZero}. These engines learn and imitate playing styles that resemble human intuition. The current  World Champion, Magnus Carlsen has a rating around 2882.
By contrast {\it Stockfish and AlphaZero} are estimated to have ratings above 3500, depending on hardware and testing conditions - that is, they are simply too strong even for the best players.}.

There are many other successful applications of machine learning.  In
particular there are the machine learning programs working in the
discrimination of credit card applications; in recognising handwritten
zipcodes; in counting small volcanoes in images of Venus; and in speech
recognition using hidden Markov models.  There are many medical applications,
for example in the diagnosis of abdominal pain which I am going to talk about
today.  There are several automated programs for establishing human genome
sequences.  There are also machine learning programs used for predicting
seismic events like earthquakes, in classification of tissue samples for
breast cancer screening, in prediction of financial indices such as exchange
rates, text categorisation and many others.

In this lecture I will discuss several important developments in the field of
machine learning and consider some prospects: what is the future of the
subject?  I will start by explaining {\bf what learning is}, then I will
consider several different inductive learning models using the {\bf Bayesian
  approach}.  After that I will move to our current interest in developing a
new type of universal learning machine called {\bf Support Vector Machine}.
Then I would like to spend some time talking about the prospects offered by
ML.  In particular I am going to draw your attention to a new and very
promising development called {\bf transductive learning}, which may allow us
to deal with previously unformalised concepts such as insight, intuition etc.

What I would also like to emphasise in this lecture is that machine learning
is not just an experimental science, nor is it just a theory of learning.
Only by engaging with both the theory and the experiments can one really make
progress in this subject.  What I aim to show is that the theory allows us to
design good systems and the experiments allow us to validate the theory.

\section{What is Learning?}

So what is learning?  According to Webster's dictionary ``to learn'' means
``to gain knowledge, or understanding, or skill, by study, or instruction, or
experience''.

I can try to represent this idea in a simple graphical form where in the
figure below there is a box called {\bf experience} or {\bf data} or {\bf set
  of examples} and a box called {\bf knowledge}.  The arrow between these two
boxes is called {\bf learning}, the process whereby we learn something out of
experience -- out of data, and we gain some new knowledge.  
Knowledge gained through learning partly obtained from a description of what
we have observed, and partly obtained by making inferences from (past)
data in order to predict (future) examples.
Obviously if data
have no regularities, any law incorporated into them, we won't be able to find
any new knowledge.  In other words, in random data, there is no knowledge to
be found.

\begin{figure}
\centering
\includegraphics[width=0.8\textwidth]{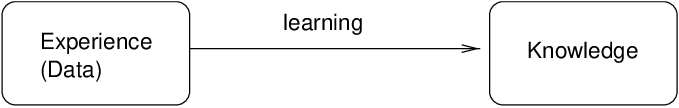}
\caption{Learning from experience}
\end{figure}

In answer to the question, what do we do?: we are trying to find some
regularity, some knowledge in data.

The next question we may ask is, why do we do this? and the answer is, of
course, we would like to make some predictions for future examples or to make
some decisions or just simply to understand.   If we would like to answer the question how, how do we do learning? the
simplest answer is, we do it by searching -- we are physically
searching for ``good'' models in our data and then we use these models for
future predictions.
There are also different types of machine learning
such as supervised/unsupervised learning or online/offline learning
but they can all be described by the same scheme.

Let me now give you an example of a real application of supervised learning~\cite{GamTh90}.  This is an application in the medical domain, in the
diagnosis of abdominal pain, and the basic set of data was collected at a
hospital in Scotland.  There were about 6,000 patient records, of patients who
suffered from abdominal pain.  Each patient had 135 binary symptoms, such as
are represented in the figure below, and there were 9 possible diagnostic
classes to which each of these patients was assigned: classes like
appendicitis (App), dyspepsia (Dys), perforated peptic ulcer (Ppu) etc.  The
task was to extract out of this information the set of relevant symptoms for
each diagnostic class -- for each disease -- so that the program would be able
to predict with certain probability the disease of a new patient.  So the
input information -- the input matrix -- looked like Table~\ref{tab1}.

\begin{table}
\begin{center}
\caption{Medical Database: Scotland; 6,387 patients; 135 symptoms;
  9 diseases}\label{tab1}
\begin{tabular}{|c|c|c|c|c|c|c|}
\hline
ID&Age&Sex&Pain-on-site&Nausea& ....&Diagnosis\\
\hline
84136&18&M&Y&N&......&App\\
65140&34&F&N&N&......&Dys\\
71853&61&M&Y&Y&......&Ppu\\
.....&...&...&...&...&......&...\\
\hline
\end{tabular}
\end{center}
\end{table}

There is an input set of symptoms -- attributes or feature vectors, or random
variables, $X \in \{x_{1}, x_{2}, ..., x_{n}\} $, and each variable has a set
of values.  The classes $C_1$, $C_2$, ..., $C$ -- or we can call them groups or
diseases, or diagnostic groups -- are all mutually exclusive and exhaustive.
So basically we are trying to find a mapping between symptoms and classes.  Our
task is to find a combination of symptoms which will indicate a certain class
with a certain probability.  An algorithm called G\&T was developed and the learning part of this algorithm is shown below.

\vspace{0.3cm}
\underline{Algorithm (G\&T)}

\begin{tabbing}
mm\=mmmmm\=mmm\=mmmmmmmmmmmmmmmmmmmmm\kill
{\em Matching-with-Selection} Learning (examples, attributes, classes)\\
\>{\bf input}:\> {\em examples}\\
\>\> {\em attributes}\\
\>\> {\em classes}\\
\>{\bf if} \> {\em examples} is empty then {\em terminate}\\
\>\> else if all examples have the same\\
\>\> class then {\bf return} {\em class}\\
\> {\bf else}\\
\>\>1. for each class $c$ calculate $\chi^2$ values for each attribute\\
\>\> and choose the attribute with the highest $\chi^2$\\
\>\>2. partition the current set of data into two subsets\\
\>\> including and excluding the selected attribute\\
\>\>3. repeat for each subset until the termination is met\\
\>\>4. calculate probability p and confidence limits\\
\>{\bf end}\\
\>{\bf output}: combination of attributes with probabilities of classes\\
\>\>and confidence limits\\
\end{tabbing}

The first step in this algorithm is to find a set of the most important
symptoms for each diagnostic class.  This is done by using a set of
statistical tests, in our case the $\chi^2$ test.  The second step is to
partition the current set of our data, our training examples, into two subsets
of {\em including} and {\em excluding} the selected attribute.  And the third
step is to repeat the algorithm for each subset until the termination
condition is met.  Once this is done we can calculate how many patients who
had a certain combination of symptoms also developed one of the diseases, and
we can also calculate how many patients had the same combination of symptoms
but didn't develop the disease.  Once we take a ratio between the two we shall
have the best estimate or a probability.

This is repeated for each diagnostic class so at the end of the day we shall
have all the combinations of symptoms important for each diagnostic class with
the corresponding probabilities.

After this we should be able to make a prediction for a new patient and
classify this new patient according to his or her symptoms.  Once we have
classified all the new patients -- it is called the testing set -- we should
be able to estimate how good the results are, by comparing the results of the
algorithm with the performance of the doctors (in percentage of correct diagnoses), and in Table~\ref{tab2}
you can see the result of this investigation.  
In the same table you can see the results of performances of several other 
algorithms re-implemented using the G\&T model. These include the ``Simple Bayes''
model, and the CART model - for details see~\cite{GamTh90}.      
The consultants are still doing
better but the performance of the programs is not that far off.

\begin{table}
\begin{center}
\caption{Performances of doctors and programs}\label{tab2}
\begin{tabular}{|l|c|}
\hline
Consultants& 76\%\\
\hline
Registrars & 65\% \\
\hline
Junior Doctors & 61\% \\
\hline
G \& T & 65\% \\
\hline
G \& T (Simple Bayes) & 74\% \\
\hline
G \& T (CART) & 64\% \\
\hline
\end{tabular}
\end{center}
\end{table}

I will use this example now in order to describe the most important
characteristics of a learning algorithm.  The first is the {\bf representation
of the data} (a matrix in our example).  Another important characteristic of
a learning algorithm is the {\bf search procedure}.  And the last one is the
{\bf performance} of the algorithm -- we need some sort of criteria for
estimating the performance of the chosen algorithm.

In addition to these three important issues there are also some important
points to decide, in particular about the {\bf learning principle} which lies
at the heart of the design of a learning algorithm.  When I say learning
principle, I mean we can use inductive learning or
a neural network model or we can use, for example, Minimum Description Length
principle~\cite{Ris84}.

But among all these important issues perhaps the most difficult problem lies
with the search for the ``good'' models.  For example, in the chess game
programs there could be up to $10^{40}$ legal positions, or if we consider
that each player can make about 50 moves in a game, and the branching factor
of the game is about 35, so the tree will have $35^{100}$ possible moves, an
unimaginable number.  In our medical example if each of the symptoms could
have only 2 values, yes and no, then with about 135 possible symptoms we would
have $2^{135}$ possible combinations.  So the question arises about the
computational efficiency of our algorithm: is it possible to build an
efficient search procedure?

This is not a new question.  In the 1930s and 1940s, work by Godel, Kleene,
Church and Turing showed that there are truths that are not deduceable, and
functions that are not computable.  It started from recursion theory and now
it is a whole branch of computer science called complexity theory.  Basically,
it can be shown that certain problems are harder to compute than others.  Many
problems can be computed in polynomial time (time which is a polynomial
function of the size of the problem), other problems can take exponential time
or even worse.  This problem of computation is called ``combinatorial
explosion'', or ``exponential growth'', or the problem of dimensionality.
Richard Bellman back in the 1960s called this the ``curse of dimensionality''.

\begin{figure}
\centering
\includegraphics[width=0.75\textwidth]{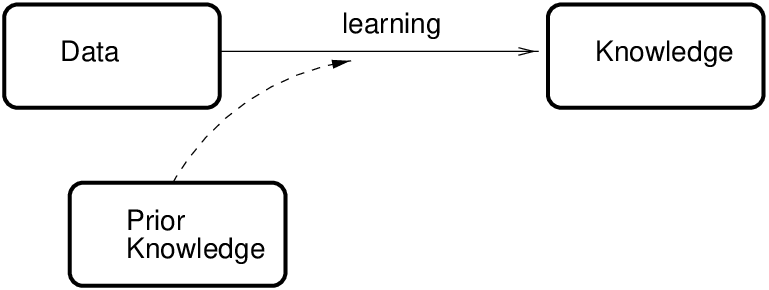}
\caption{Solution 1}
\label{solutions_1}
\end{figure}

So what I would like to present now are some possible solutions to this
problem.  In particular I am going to talk about two possible solutions.  As
you can see in Figure~\ref{solutions_1}, we have the data, the 
information processing act called learning and then the box called knowledge,
but in addition to this I can always try to use, somehow, some prior knowledge
-- prior information, and therefore to reduce the amount of calculation.

\begin{figure}
\centering
\includegraphics[width=0.75\textwidth]{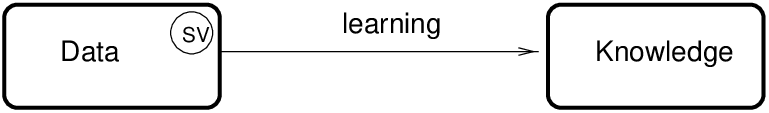}
\caption{Solution 2}
\label{solutions_2}
\end{figure}

Another way of tackling the problem, of reducing the amount of computation, is
just to reduce the original set of data - our experience - to a very
essential set of examples, called support vectors, and then use just that
subset of data in order to learn and gain new knowledge - see Figure \ref{solutions_2}.

In the first approach the prior information could be a set of additional
assumptions such as the assumption of conditional independence or the
assumption that all our examples are distributed identically and
independently.  And in the second approach, when we use only essential information, for example, support
vectors, it means that somehow we find a mechanism to compress the data up to
the only ``important'' set of vectors in order to solve our problem.

\section{Bayesian Approach}

\begin{figure}
\centering
	\includegraphics[width=0.6\textwidth]{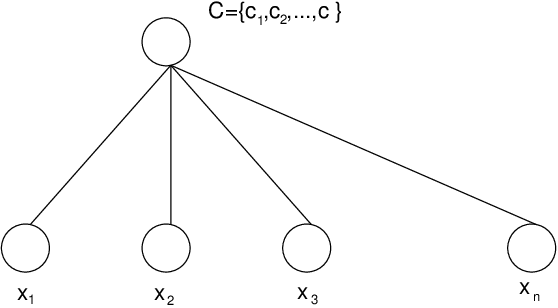}
\caption{Simple Bayes model}
\label{simple_bayes}
\end{figure}

Let me now show you how the first approach works, when we use prior
information - prior knowledge.  The first model I would like to show you is
the so-called ``simple'' Bayes model when we can make an assumption of
independence, that is, all our symptoms $X_1$, $X_2$, up to $X_n$ are
independent given the disease, and this can be represented in graphical form
where on the top level is our node with all possible diagnostic groups or
possible diseases and on the next level down are all possible symptoms, so we
assume that each of the symptoms $X_1$, $X_2$ is independent given the
disease.  This is represented graphically in Figure \ref{simple_bayes}.
The performance of ``simple'' Bayes model in our medical example is shown
in Table 2.

We can expand this idea further and make a graphical representation
where not all symptoms are independent but only some of them (the nodes that
are not connected are assumed to be conditionally independent). If we represent
this idea using a graph, and if we also supply a set of conditional
probabilities to this graph, then it is called a Bayesian Belief network
(BBN).  By definition, a Bayesian belief network is a directed acyclic (no
loops) graph, and a set of conditional probabilities on that graph.

\begin{figure}
\centering
\includegraphics[width=0.6\textwidth]{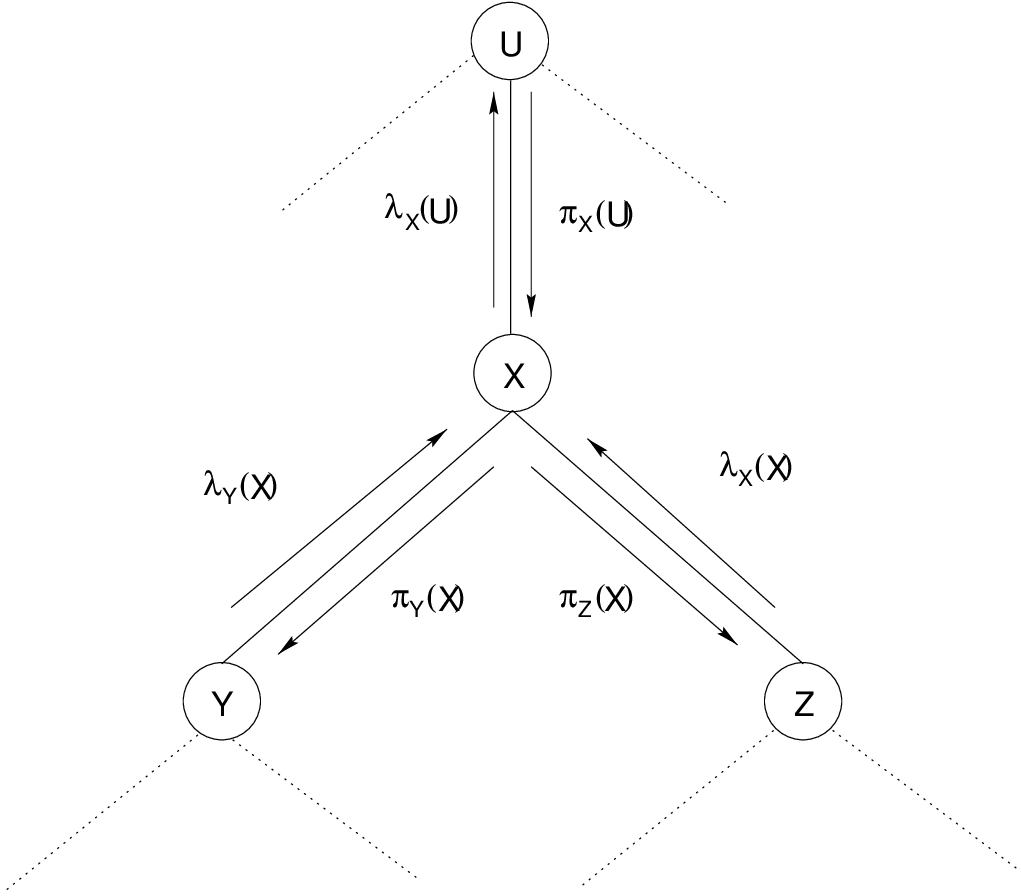}
\caption{Local computations on a tree structure}
\label{pearl}
\end{figure}

In 1982 Pearl developed an algorithm~\cite{Pearl} for a tree structure where he
considered only parents and children for each node on the tree.  If you look
in Figure~\ref{pearl} you can see that node X has one parent U and two children Y and
Z.  If an observation is made then we would have to revise our original
probabilities, and calculate posterior probabilities for each node.  What
Pearl suggested was, let's consider information coming from children as {\em
  lamda} messages or likelihoods, and information coming from a parent as a
{\em pi} message or prior probability, and fuse this information in node X.
These are local computations because we consider only a node with its parents
and its children and we are not considering nodes which are widely separated
on the tree.  And if we do this procedure step by step down the tree, we can
build up and calculate the overall structure

The same idea has been generalised for a general graphical structure, as shown
below, and by decomposing the graph into small groups called cliques
and making a clique tree; this allows us to calculate the posterior belief or
posterior probabilities as soon as we make an observation.  Behind this method
lies the method called Gibbs potentials~\cite{LaurSpieg} and it guarantees that we always calculate our results correctly. 

\begin{figure}
\centering
	\includegraphics[width=0.6\textwidth]{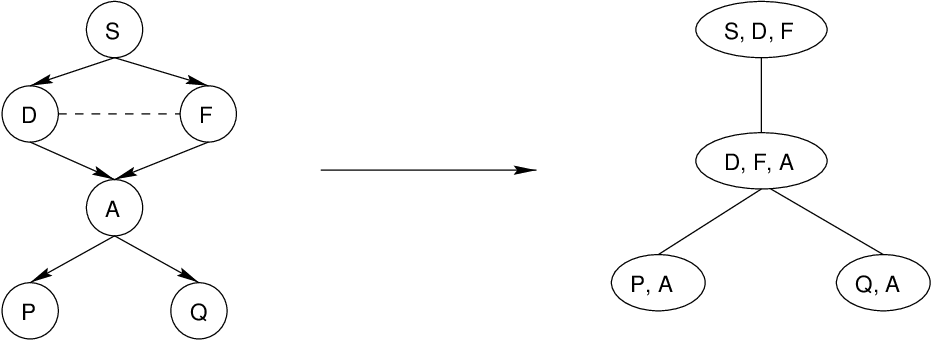}
\caption{Local computations on a graphical structure}
\end{figure}

\begin{figure}
\centering
\includegraphics[width=0.8\textwidth]{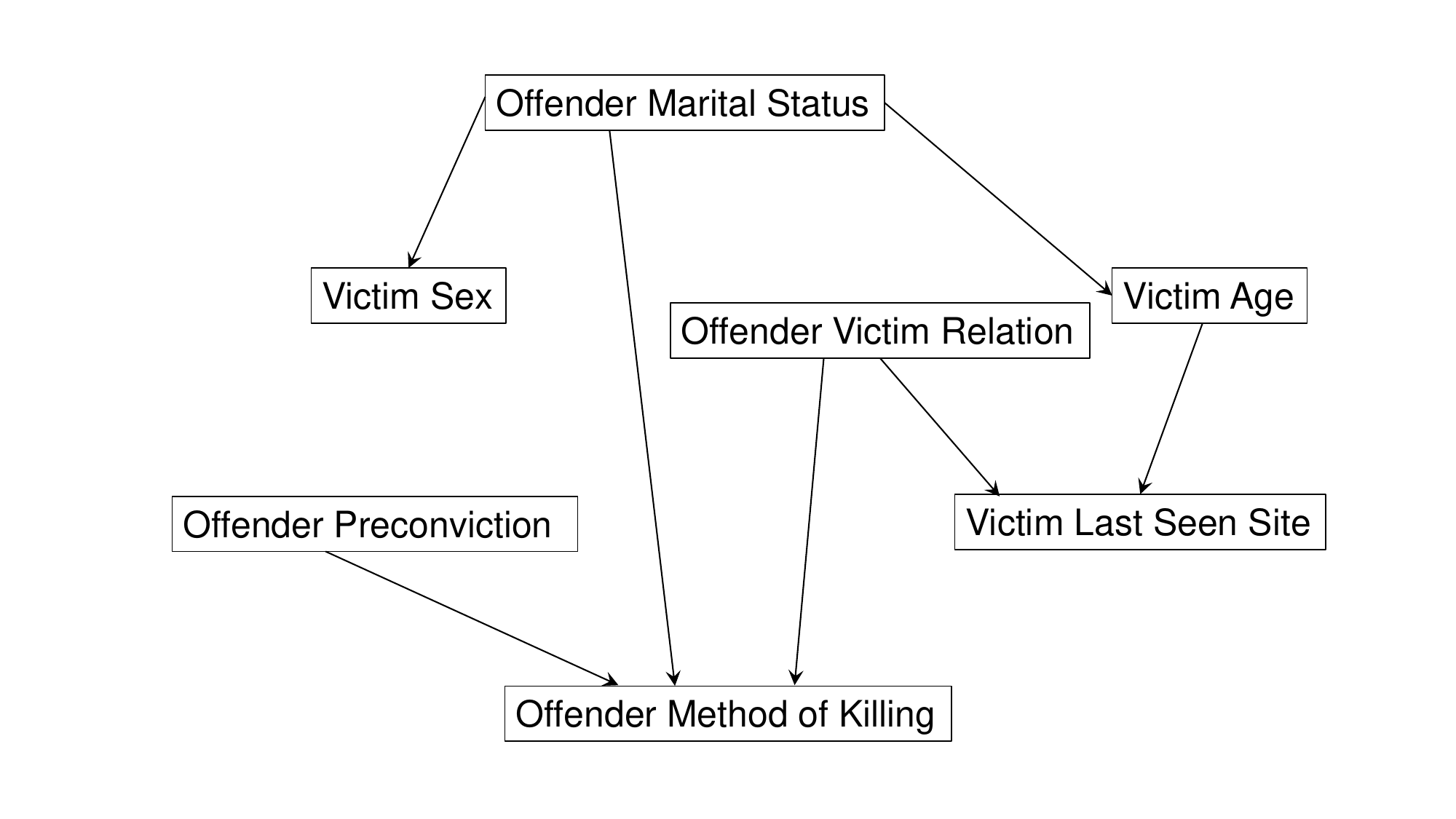}
\caption{BBN for offender profiling}\label{offender_profile_bbn}
\end{figure}

A number of computational systems have been developed using these ideas.  In
1987 we developed a Causal Probabilistic Reasoning System which
was later (1992) redeveloped into a shell system called PRESS ~\cite{GamLuo95}.  PRESS has a number of important and useful options
that allow a user to deal with a mixture of discrete and continuous variables,
to construct a BBN out of data (structural learning), to determine the most
important variables, etc.  The systems have been tried in a number of
applications and one of the most recent, shown in Figure~\ref{offender_profile_bbn}, is Offender Profiling -- a collaborative project
with the Home Office~\cite{AitGam96}.
The aim of the project was to develop statistical profiles of offenders using some
data and knowledge held by the detectives. The data set
was initiated by Derbyshire Constabulary in 1986,
and there were 320 cases of offenders described by 5 attributes such as offender
preconviction, offender age, etc, and 8 attributes related
to the victims and scenes of crimes.
Several Bayesian Belief networks have been
created by the detectives and one of them is shown here, with some of the corresponding
probabilities in the table below.  Then as soon as information arrives (an
observation is made) about a particular node -- victim age etc. -- this
information, the probabilities can be revised and we can make some
predictions - see Table~\ref{tab3} - about the offender's preconviction, marital status, etc.

\begin{table}
\begin{center}
\caption{Conditional probabilities for offender profiling}\label{tab3}
\begin{tabular}{|lccc|}
\hline
\multicolumn{4}{|p{100mm}|}{Probabilities for offender characteristics for a female victim, aged 0-7 years, found strangled outside her own home}\\
\hline
Characteristic & Outcome & \multicolumn{2}{c|}{Probability}\\
&& Initial & Revised \\
\hline
Living with partner & Yes & 0.24 & 0.36\\
&No & 0.76 & 0.64 \\
Relationship & Known & 0.57 & 0.11 \\
& Unknown & 0.43 & 0.89 \\
Preconviction & Yes & 0.73 & 0.70 \\
& No & 0.27 & 0.30 \\
\hline
\end{tabular}
\end{center}
\end{table}

I can summarise this approach by saying that we use additional prior
information in the form of either making some assumptions of independence in
the simple Bayes model or using Bayesian belief networks to simplify the
computational procedure and make it much more efficient.  Obviously, the gain
in computational efficiency is just one of several benefits in using
additional (prior) information.  In general, the use of prior knowledge allows
us to develop very powerful algorithms.  For example, here is a
general Bayesian algorithm developed by V. Vovk~\cite{Vovk90}. Suppose that {\em a
  priori} we have a set of possible hypotheses: $H_1$, $H_2$, ....; more
generally we can have some family of hypotheses $\{H_\theta / \theta \epsilon
\Theta \}$ -- it can be finite, or countable, or continuous, and instead of finding the
best hypothesis, we compute the ``weights'' for all the hypotheses
($H_\theta$).  These weights would reflect how much we trust (or like) each of
our hypotheses.  We start by assuming the prior weights $P(d\theta)$, and when
a new example arrives, every $H_\theta$ gives a prediction $P_\theta$.  Then
the Aggregating Algorithm merges all $P_\theta$ with accordance of their
weights $P(d\theta)$.  When the actual hypothesis is
disclosed we compute the loss $l_\theta$ supported by 
hypothesis $H_\theta$, and then recompute the
weights:

$$ P(d\theta):= \frac{e^{-\eta l_\theta}P(d\theta)}{\int e^{-\eta l_\theta}P(d\theta)}$$

\noindent where $\eta$ is a learning rate.  As one can see if a hypothesis is
not ``correct'' (after disclosure of the real class) - this is reflected in
a big value of loss function $l_\theta$ - then 
the weights are slashed.

Special cases of this algorithm are: Bayesian merging and weight updating,
the weighted majority algorithm in pattern recognition, Cover's universal
portfolios algorithm and many others.

\section{Support Vector Machine}

Another way, as I have already mentioned, is just to compress the data itself
and make use of only an ``essential'' set of examples - support vectors.
This approach was originally developed by V. Vapnik~\cite{Vapnik1995}, who is now
leading this research 
in our Department.

Let me start again from the familiar picture of the learning 
process (Figure 4). We shall follow our scheme again: from data to new knowledge through inductive learning.

The {\bf data}
are represented with a {\bf set of examples} $(x_{1}, c_{1}), .... , (x_{l},
c_{l})$;  $ x_{i} $ is a vector of attributes and $ c_{i} $ is a class which
takes value either positive (+1) or negative (-1).  The task is to find a
decision rule that separates examples into positive (+1) and negative (-1)
classes.  The {\bf rule} would then represent the {\bf knowledge}.  This is a
typical pattern recognition problem.

The main idea is to map our original set of vectors into high dimensional
feature space, and then to construct in this space a linear decision rule (an
``optimal hyperplane'').  Then we can find the maximal margin between the
vectors of the two classes.  The vectors that determine the margin are called
``{\bf support vectors}'', and they can replace the whole training set of
examples.

\begin{figure}
\centering
\includegraphics[width=0.7\textwidth]{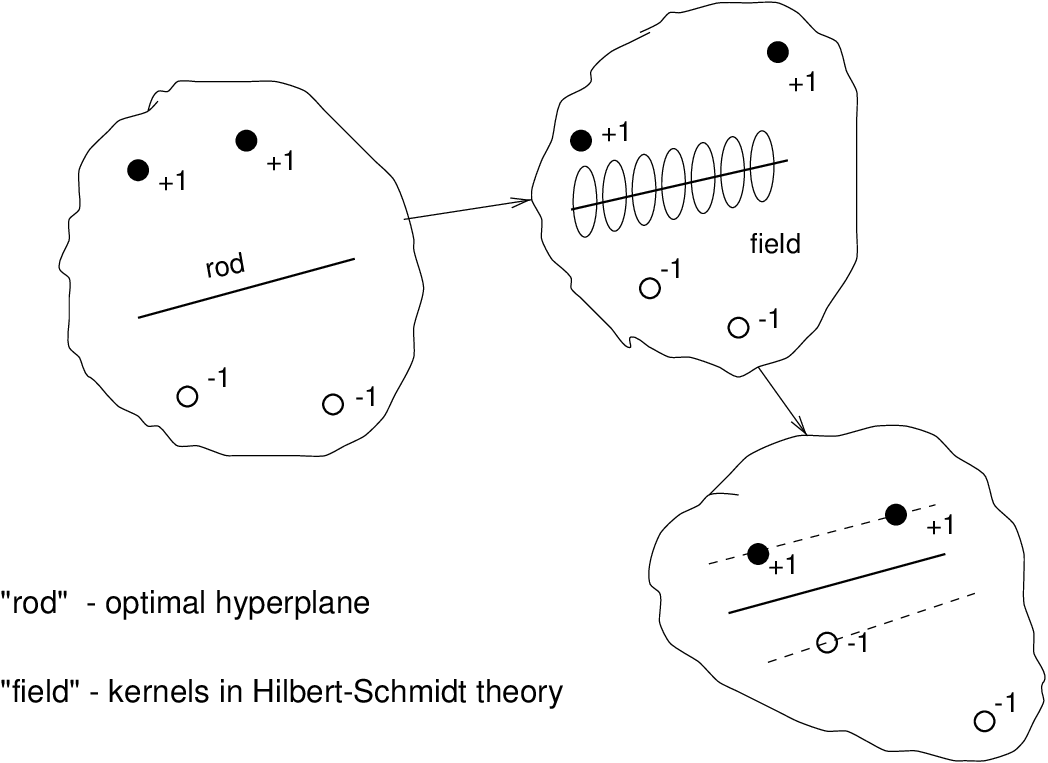}
\caption{Support Vector Machine for linear separable case}
\label{inaugs6a}
\end{figure}

Let me use a physical analogy to explain the main ideas.  
I would represent a decision rule with a magnetic rod.  This rod
can move freely in the space between positive and negative examples, and it
generates a field around itself with sharp boundaries.  But as soon 
as the field reaches several nearest
negative and positive examples, it stops propagating, and the rod will be
fixed at this point.  Those negative and positive examples that are on the
boundary of the field are called the ``support vectors''.  We
can now make predictions by determining on which side of the margins the new
examples come. The idea is illustrated in Figure~\ref{inaugs6a} where black
dots represent positive (+1) examples, and white dots represent negative (-1)
examples.

This is an exact description of the linear and separable case.  In general we
could present our ``field'' as the {\bf kernel functions} $K(x, x^\prime)$ of the
Hilbert-Schmidt theory.

The training procedure in this algorithm, to find the optimal hyperplane,
amounts to solving a constrained quadratic optimisation problem.  The solution
of this problem is a unique global minimum.

One can also assess the performance of the algorithm, and it has been shown
 that the probability of error is bounded by the ratio of the
expected number of support vectors to the size of the training set:

$$Pr(\mbox{err}) \leq \frac{E(\#SV)}{\#\mbox{training\_examples}} $$

This is a constructive implementation of the main idea, and it boils down to
the compression of the original set of data to ``only'' the support vectors
set (usually 3\% --- 5\% of the original set).

The next question we may ask is, why are the SV machines so successful in
compressing the data, where does their power of generalisation lie?  The
answer to this question turns out to be a feature of the problem called {\bf
  VC-dimension} (Vapnik-Chervonenkis).  One of the main results of learning
theory has been the discovery of the following inequality that holds with
probability at least $1 - \eta$ and gives
the bound for probability of errors in the testing set:

$$ Pr(\mbox{err\_testset}) \leq Pr(\mbox{err\_trainset}) + \Phi(\frac{l}{h},  \frac{-\ln\eta}{l})
$$

\noindent where $h$ is the VC-dimension of the set of functions, and $l$ is
the number of examples in the training set.

Recall our example when the field becomes more powerful, and the band becomes
wider and wider, and the rod gets more and more restricted.  It turns out that
the VC-dimension of the set of possible rod positions becomes smaller and
smaller.  And when at the limit we reach a point where we cannot move the rod
any more, the VC-dimension becomes minimum, and thus, the SV-machine
implements the principle of minimizing the VC-dimension.  V. Vapnik obtained
some accurate estimates of the VC-dimension of this set depending on the width
of the band.

The idea of learning using SV-machines is currently being tried in several
application areas: speech recognition, faces recognition, tomography, medical
diagnosis and it is already clear that this new approach has very strong
generalisation abilities and predictive power.


\section{Transduction}

The last point I would like to touch on in this lecture is a different type of
learning.  So far we have been discussing mainly inductive learning.

Let me now reformulate the problem and ask why do we need the ``knowledge'' at
all?  One possible answer to this is that we want to say something about (to
predict) future examples.  So, we can say that we use ``knowledge'' to predict
``new examples'':

\begin{figure}
\centering
\includegraphics[width=0.7\textwidth]{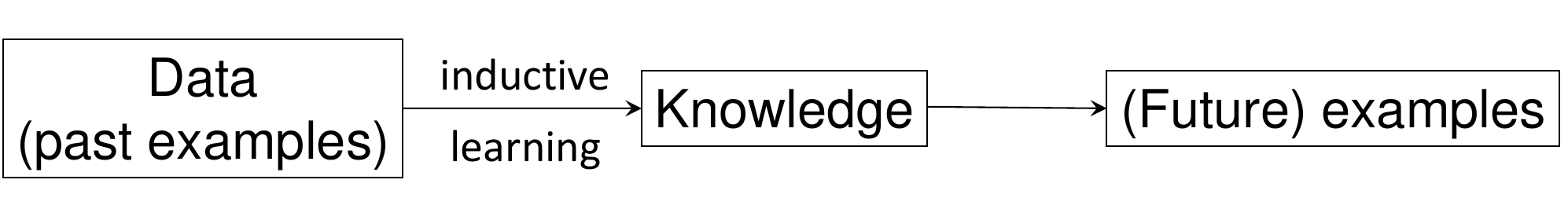}
\caption{Prediction in inductive learning}\label{fig11}
\end{figure}

The ``knowledge'' in our last example was a decision rule or optimal
hyperplane.  Obviously we spent a great deal of (computational) effort in
order to construct the rule and to find the support vectors.  When the rule is
constructed it of course allows us to predict the classes (positive or
negative) of new examples.  In fact, we can then predict any new examples,
even an infinite number of new examples can be classified.  In reality,
however, we are interested only in predicting a finite number of new examples.
If we require to predict {\em only some} new examples, but have already {\em
 knowledge} to predict {\em any} new examples, then, perhaps, we have
``overkilled'' the problem.

Then, the interesting question arises: do we always need to go from the
particular (examples -- the data) to the general (knowledge) -- this is
usually called induction, and then back to the particular (future) examples --
this is usually called deduction?

\begin{figure}
\centering
\includegraphics[width=0.7\textwidth]{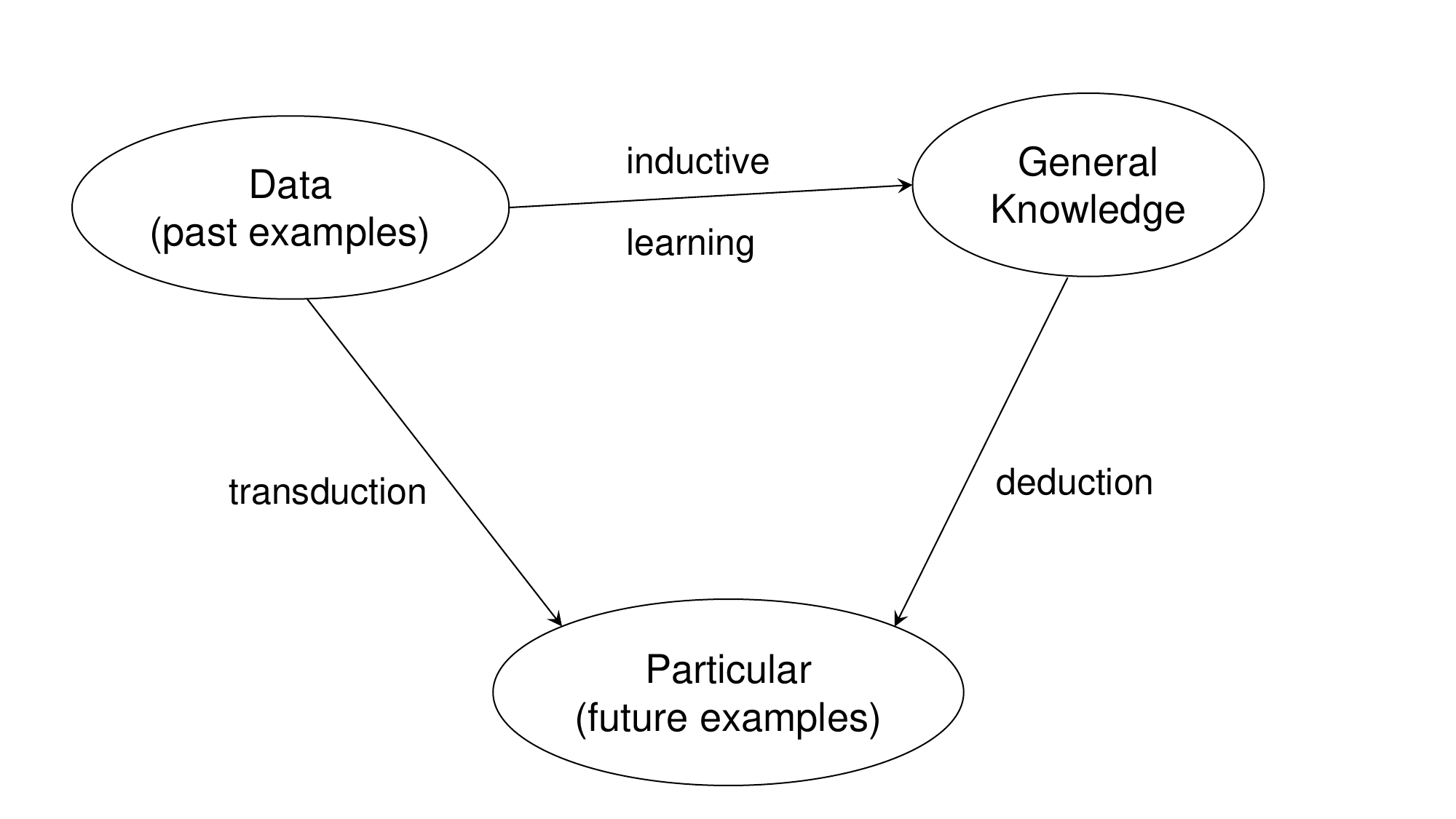}
\caption{Prediction in transductive learning}
\label{transduction}
\end{figure}

Can we make a short cut, and go from particular directly to particular?  We shall call
this short cut a {\bf transduction}~\cite{Vapnik1995}.  It looks as if this way we shall
make some ``savings'' -- instead of having ``general knowledge'' it is more
efficient to get ``specific knowledge'' about particular instances.  So, the
transduction is going from particular (past) examples to particular (future)
examples without any attempt to ``generalise'' our experience and get
``general knowledge'' - see Figure~\ref{transduction} (taken from~\cite{Vapnik1995}).

We hope that transduction will (1) allow us to make the learning
process computationally more efficient; (2) facilitate some theoretical
advance, since it will be easier to prove some properties of a learning
algorithm (inferring from particular to particular).

Let me illustrate this last point with an example.  Recall our picture of
positive and negative examples (black and white dots) separated by the optimal
hyperplane (figure \ref{inaugs6a}).  
Perhaps one of the disadvantages of this induction learning
is that this is a ``black and white'' picture, with no ``grey'' area.

When we found our separating surface (decision rule) we obtained some ``black
and white'' picture; we have black points, and white points, and we classify
all points of one side of the margin as black, and on the other side of the
margin as white.  What we would like to have is some tints of ``grey'' colour,
expressing our {\em confidence} - see Figure~\ref{numbers}.

For example, if a new point is near to the black points area, we expect it to
be a black point, with significant confidence.  If it is near to the white points,
we are confident that it should be a white point.  And if it is near the
separation surface, we are not sure, this is a grey region.

Here is another example: a clear digit 2 (could correspond to the ``black''
area), clear digit 7 -- ``not 2'' -- corresponds to the ``white'' area, and
something that is not clear -- ``grey'' area:

\begin{figure}
\centering
\includegraphics[width=0.5\textwidth]{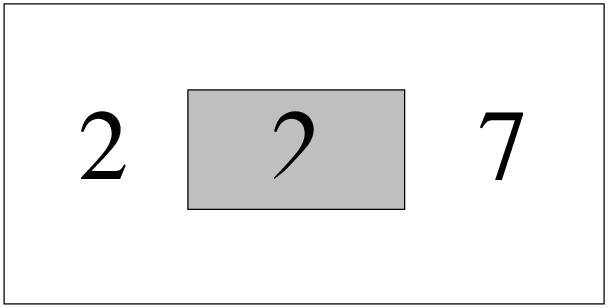}
\caption{Uncertainty quantification by {\it confidence} in digit recognition problem}
\label{numbers}
\end{figure}

\begin{figure}[h]
\centering
\includegraphics[width=0.6\textwidth]{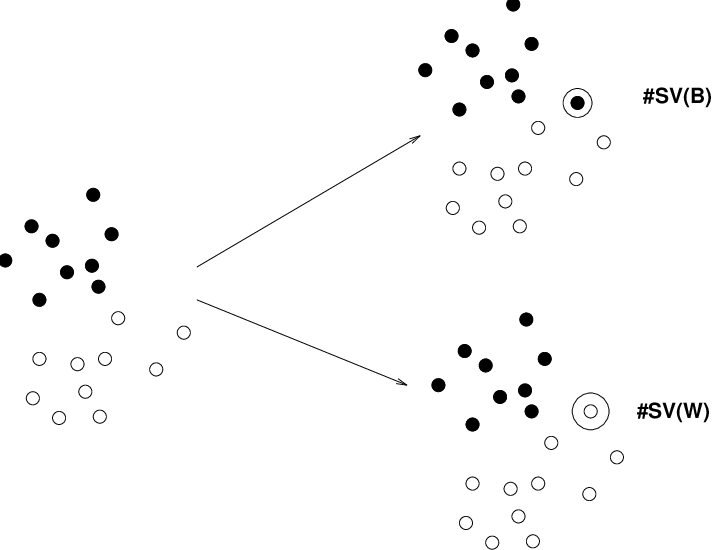}
\caption{Prediction with confidence}
\label{prediction}
\end{figure}

With induction, no efficient procedure for finding this ``grey'' picture is
known.  Let me now explain how this idea of confidence can be expressed using
transduction~\cite{VovkGam}.  Assume that we now have our training set of data and also a new
example which has not been classified yet.  We can represent it with the
following picture - see Figure~\ref{prediction}.

In the picture $\#SV(B)$ is the number of support vectors in the black
picture, and $\#SV(W)$ is the number in white.  We now have two pictures: in
one the new example is classified as ``black'', and in the other, it is
classified as ``white''.  Recall that there is no general separating curve any
longer, because there is no ``general knowledge'' in this transduction
process.  All we want is to predict whether the new example is white, or
whether it is black, i.e. to make a prediction for that particular point.  All
we can say is that this new example is a support vector in at least
one of the two pictures.  
Then, we can formulate a simple rule that classifies our
new example with a certain confidence:

\vspace{2mm}
\parbox{80mm}{
Prediction rule ($l$ is the number of examples and SV(B) and SV(W) are the support 
vectors in the black and white pictures respectively):

\vspace{2mm}

{(1) WHITE if
$$ \mbox{new\_example} \in SV(B) \&  \mbox{new\_example} \not\in SV(W) $$
or
$$  \mbox{new\_example} \in SV(B)\cap SV(W) \& [\#SV(B)<\#SV(W)] $$
CONFIDENCE :  $ 1 - \frac{\#SV(B)}{l} $

\vspace{2mm}

(2) BLACK if
$$   \mbox{new\_example} \in SV(W) \&   \mbox{new\_example} \not\in SV(B) $$
or
$$  \mbox{new\_example} \in SV(B)\cap SV(W) \& [\#SV(B)>\#SV(W)] $$
CONFIDENCE :  $ 1 - \frac{\#SV(W)}{l} $

\vspace{2mm}

(3) NO PREDICTION
$$   \mbox{new\_example} \in SV(W)\cap SV(B) \& [\#SV(B)=\#SV(W)] $$

\vspace{2mm}

NO CONFIDENCE
(any prediction with no confidence)}}

\vspace{3mm}

This prediction rule works well, and full confidence is expressed with 1, and
no confidence with 0.  For example, if a {\it new example} is a support vector
in the ``black'' picture, and it is not in the ``white'' picture, we classify
it as white with the confidence $ 1 - \frac{\#SV(B)}{l} $; if the fraction of
support vectors in the ``black'' picture is small, then the confidence is high
(close to 1).

\medskip
{\it Remark 1}. These ideas were discussed extensively, and this led us to the development of {\it conformal predictors}~\cite{VovkGam,TCJ,ICML99,IJCAI}. The method is ideally suitable as a framework for uncertainty quantification; the most definitive exposition of Conformal Predictors theory is given in~\cite{books}.

\medskip
{\it Remark 2}. Recent development of {\it Deep Learning} and {\it Large Language Models} (LLMs) has
a considerable impact on the conformal predictors. CPs have evolved to address the needs of these powerful but often overconfident and lack calibrated uncertainty estimates~\cite{Patrizio}. By enhancing
calibration, interpretability, and trust, Conformal Predictors play an increasingly critical role in making modern machine learning systems
more reliable and statistically valid. 

One of the weaknesses of LLMs is that they do not have a notion of the ground truth; there is a hope that CPs could bring this notion to the large language models.

\section{Conclusion}

Let me now return to our original picture of inductive learning and claim that
following the success of 20th century science (e.g. quantum mechanics), true
knowledge is inaccessible and the best we can hope for is to substitute for it
the best hypothesis.  But even the best hypothesis is very often difficult to
find, and this could lead to the position of agnosticism.  However, I will
abandon this line of argument for now since it will take us into a new and
deep philosophical discussion about the nature of knowledge.

There is one last comment I would like to make in conclusion.  I started this
lecture by saying that learning is intimately connected with intelligence.
For centuries people have argued about what lies at the basis of human
intelligence.  Is it our logical abilities and generalisation skills that lead
us to gain new information and discover laws of nature?  Or is it human
intuition and insight that play the most important role in different
discoveries as well as in the huge area of human intelligence called arts and
humanities?  It is clear that logical ways of thinking and generalisation
abilities can be connected with deduction and induction, while transduction
can be a way to formalise intuition and insight.  This opens a fascinating
prospect for future research but perhaps this is a subject for future
lectures.

\bibliographystyle{splncs04}

\end{document}